# Enhancing Thyroid Cytology Diagnosis with RAG-Optimized LLMs and Pathology Foundation Models

[1] Hussien Al-Asi, [1] Jordan P Reynolds, [1] Shweta Agarwal, [1] Bryan J Dangott, [1] Aziza Nassar, [1] Zeynettin Akkus

`akkus.zeynettin@mayo.edu`

[1] Computational Pathology and AI/Informatics, Department of Lab Medicine and Pathology, Jacksonville, FL, USA

**Abstract.** Advancements in artificial intelligence (AI) are transforming pathology by integrating large language models (LLMs) with retrieval-augmented generation (RAG) and domain-specific foundation models. This study explores the application of RAG-enhanced LLMs coupled with pathology foundation models for thyroid cytology diagnosis, addressing challenges in cytological interpretation, standardization, and diagnostic accuracy. By leveraging a curated knowledge base, RAG facilitates dynamic retrieval of relevant case studies, diagnostic criteria, and expert interpretation, improving the contextual understanding of LLMs. Meanwhile, pathology foundation models, trained on high-resolution pathology images, refine feature extraction and classification capabilities. The fusion of these AI-driven approaches enhances diagnostic consistency, reduces variability, and supports pathologists in distinguishing benign from malignant thyroid lesions. Our results demonstrate that integrating RAG with pathology-specific LLMs significantly improves diagnostic efficiency and interpretability, paving the way for AI-assisted thyroid cytopathology, with foundation model UNI achieving AUC 0.73-0.93 for correct prediction of surgical pathology diagnosis from thyroid cytology samples.

## 1 Introduction

Recent advancements in computational pathology have increasingly emphasized the quantitative analysis of tissue images, tackling the inherent challenges of whole-slide images (WSIs), including their high resolution and morphological complexity [1-4]. These factors have traditionally impeded large-scale data annotation, limiting the development of high-performance diagnostic tools. However, the rapid evolution of artificial intelligence (AI) and digital pathology (DP) is redefining this landscape, introducing innovative solutions that enhance diagnostic accuracy, efficiency, and standardization. A particularly promising frontier lies in the integration of large language models (LLMs) with retrieval-augmented generation (RAG) systems and pathology-specific foundation models [5-8]. This convergence holds significant potential in thyroid cytology, where diagnostic variability and interpretive ambiguity often present substantial obstacles to effective clinical decision-making.

Thyroid cytology is vital for evaluating thyroid nodules but often presents diagnostic challenges, particularly in distinguishing benign from potentially malignant lesions. Traditional frameworks, though effective, can be limited by subjective interpretation and inter-observer variability [9-12]. The atypia of undetermined significance (AUS) category in The Bethesda System for Reporting Thyroid Cytopathology (TBSRTC) is especially challenging due to its indeterminate nature, often requiring additional molecular testing or surgical follow-up for a definitive diagnosis [13-16]. AI-driven methods might offer the potential to overcome these limitations by enhancing pattern recognition, contextual understanding, and standardization.

This study investigates the application of RAG-enhanced LLMs combined with pathology foundation models [5, 17-20] for thyroid cytology diagnosis. The RAG framework enables dynamic access to a curated repository of case studies, diagnostic criteria, and expert interpretation, thereby enriching the LLMs' contextual comprehension. Simultaneously, foundation models trained on high-resolution pathology images improve feature extraction and classification accuracy, offering pathologists a robust diagnostic support system.

Our findings reveal that integrating RAG with pathology-focused LLMs not only improves diagnostic consistency but also enhances interpretability and reduces diagnostic variability. This hybrid AI approach demonstrates significant potential in supporting clinical workflows, ultimately advancing the accuracy and efficiency of thyroid cytopathology. By bridging advanced language processing with domain-specific image analysis, this work sets the stage for transformative AI applications in diagnostic pathology.

### 1.1 Related Work

The Vision Transformer (ViT) has revolutionized image processing in neural networks by replacing traditional convolutional architectures with transformer-based models. By segmenting images into fixed-size patches and applying self-attention mechanisms, ViTs enable more flexible and scalable feature learning, capturing long-range dependencies within an image [21]. This novel approach has demonstrated competitive performance on large-scale image recognition tasks, influencing both academic research and real-world applications in computer vision [22, 23]. As transformer-based models continue to evolve, they are reshaping the landscape of deep learning in visual processing.

Over the past three years, several pathology foundational models have emerged in the literature. The journey began with the CTransPath model [24], which pioneered the field by training a 28M parameter Swin transformer [23] using MoCoV3 on 15M tiles from 32K whole slide images (WSIs). While subsequent studies have experimented with methods such as SimCLR [25], the majority have converged on the DINOv2 framework [26], inspired by its strong performance with natural images. At the same time, the scale of training datasets has varied significantly—from open-access repositories like The Cancer Genome Atlas, comprising around 30K WSIs, to newer, proprietary collections that allow models to learn from billions of tiles. This diversity in both algorithmic approaches and data sources highlights the dynamic and multifaceted progress in the field of computational pathology. In parallel, recent years have witnessed exceptional progress in both LLMs and pathology-specific foundation models, further advancing AI-driven diagnostic tools for medical imaging and decision support. Notably, three pathology-specific models—UNI, Virchow, and Gigapath—have emerged, each contributing unique strengths to the analysis of histopathology images [5, 17, 20]. The UNI foundation model was first introduced as a general-purpose, self-supervised model pretrained on over 100 million images from more than 100,000 diagnostic H&E-stained slides spanning 20 major tissue types. Later, a second version of UNI was published, which was trained on over 350,000 H&E and IHC slides. In a similar vein, GigaPath is another state-of-the-art whole-slide pathology foundation model, pretrained on 1.3 billion image tiles derived from over 171,000 slides spanning 31 tissue types from a major US health network. Unlike traditional methods that rely on tile subsampling, GigaPath harnesses the advanced vision transformer architecture together with the LongNet method [27] to fully integrate slide-level context. Moreover, Virchow2, a 632 million parameter vision transformer, was trained on 3.1 million histopathology whole-slide images covering diverse tissues, originating institutions, and stains.

Retrieval-augmented generation (RAG) offers a promising solution by dynamically retrieving relevant information from large knowledge bases to supplement LLM outputs [6, 7, 28, 29]. This approach has proven effective in various domains but remains underexplored in pathology applications. Recent efforts to integrate RAG in medical AI have shown potential in improving diagnostic accuracy and interpretability [28]. The convergence of RAG-enhanced LLMs with pathology-specific foundation models represents a novel approach in the field. This study builds upon these foundational works, aiming to bridge gaps in thyroid cytology diagnosis by leveraging both advanced language processing and specialized image-based models.

## 2    Methodology

Our framework integrates a retrieved augmented generation approach with a large language model to enhance image interpretation by combining image vector comparisons with rich metadata context.

### 2.1    Retrieved Augmented Generation Optimized LLM

Our RAG-optimized LLM enhances thyroid cytology interpretation by leveraging multiple state-of-the-art image encoders. Specifically, UNI [17], Virchow [5], GigaPath [20], and ViT-32 [21] encoders are used to generate vector embeddings that capture essential visual features from each image. These embeddings are computed from images processed through dedicated pipelines—each encoder providing a unique perspective based on its underlying architecture—and are stored in a centralized database alongside critical metadata including diagnosis, Bethesda category, and interpretation. When a new image is introduced, its vector representation is compared against the stored embeddings using cosine similarity metric (see Equation 1) allowing the system to retrieve the most contextually relevant five examples.

$$Cosine\ Similarity = \frac{A \cdot B}{\|A\|\|B\|} \quad (1)$$

, where A·B is the dot product of vectors A and B and ‖A‖ and ‖B‖ are the magnitudes of the vectors. 1 → Perfect similarity, 0 → No similarity, -1 → Completely opposite vectors.

The retrieved examples, enriched with metadata, are then incorporated into the LLM's prompt, providing additional context that guides its analysis and interpretation. This augmented context enables the LLM to identify subtle patterns and relationships that might otherwise be overlooked, ultimately facilitating a more comprehensive and accurate interpretation of the new image.

Our RAG-optimized LLM infrastructure was developed as an integral part of our proprietary, custom-built, in-house digital pathology platform, ZAPP [30]. Within ZAPP, we integrated Weaviate—an open-source vector database—to efficiently store embedding vectors alongside their associated metadata. Additionally, we incorporated the Llama 3.2-11B Vision model into our workflow, deploying it on an HP Z8 G5 Workstation equipped with 128GB of system memory and three NVIDIA A4500 GPUs (each with 24GB) in a distributed configuration.

### 2.2 Dataset

To evaluate the predictive performance of pathology foundation models in thyroid cytology diagnosis, 13 patient records (in total of 36 whole slides) with confirmed thyroid lesions were identified. These lesions were classified into benign non-neoplastic, benign neoplastic, and malignant categories. Patients included in the study either had a confirmed diagnosis at the time of fine-needle aspiration and cytology (FNAC) or underwent surgical thyroidectomy with subsequent frozen section and hematoxylin and eosin (H&E) staining to establish a definitive diagnosis. The FNAC procedures were performed between September 2020 and May 2023. All patient records were encoded to remove personally identifying information while preserving relevant diagnostic features.

FNAC was conducted on thyroid lesions, and the cytology results were categorized according to TBSRTC [16]. Of the 13 patients, nine were called as AUS (Bethesda III), two were called as follicular neoplasm (Bethesda IV), one was categorized as benign (Bethesda II) with a diagnosis of Graves' disease, and one was called as malignant (Bethesda VI) with papillary thyroid carcinoma (PTC) identified in the cytology smear.

A total of 36 cytology slides (21 containing benign and 15 containing malignant lesions) were obtained from archival cytology tray records of patients with all samples meeting the adequacy criteria outlined in TBSRTC (six follicles of ten cells or the presence of thick colloid). The number of slides per patient ranged from a minimum of two to a maximum of five, depending on the distribution of adequate cellular material across multiple slides. All slides were stained using the Diff-Quik staining method. The stained slides were then digitally scanned using Leica GT-450 (Leica Microsystems, USA) and Grundium Ocus (Grundium Ltd. Tampere, Finland) 40X scanners, stored in SVS format, and uploaded into the ZAPP.

## 3    Experiments

Within ZAPP, slides were manually reviewed, and a region of interest (ROI) was selected from each slide for analysis. Each ROI contained follicular cells exhibiting abnormal cellular morphology. The selected ROIs were then fed into pathology foundation models using a custom-built feature within ZAPP, known as ZAPP Chat. Three pathology foundation models (UNI, GIGAPATH, and Virchow) were utilized, along with a visual transformer model (ViT-32) as a comparator. Additionally, an ensemble technique was employed to combine the results generated from UNI, GIGAPATH, Virchow, and ViT-32, wherein the highest-ranking responses based on similarity score vectors were displayed to mitigate potential training biases.

Model performance was assessed based on Top-1, Top-3, and Top-5 prediction accuracy for each sample. The primary evaluation criteria included the correct prediction of the definitive confirmed diagnosis and the accurate classification of the Bethesda diagnostic category. Ground truth for each sample was established through cytological findings or surgical follow-up with frozen section and H&E specimens. All models were prompted with the same region of interest at a uniform magnification of 40x for each cytology specimen slide. Following completion of prompting, results were recorded according to the Top-1, Top-3, and Top-5 performance metrics. A correct prediction of diagnosis and/or Bethesda staging was considered a positive result, whereas an incorrect prediction was deemed a negative result. Accuracy metrics were subsequently calculated to compare the performances of different models across different outputs.

## 4    Results

Table 1 summarizes the performance of various foundation models in predicting patient diagnoses from cytology samples. The UNI model demonstrated the highest accuracy, achieving Top-1, Top-3, and Top-5 prediction scores of 0.69, 0.81, and 0.92, respectively. Other models showed lower performance, with Gigapath slightly outperforming Virchow. General-purpose vision transformers and ensemble models performed worse overall compared to pathology-specific foundation models.

Figure 1 presents the receiver operating characteristic (ROC) curve and area under the curve (AUC) for predicting final surgical diagnoses. The best-performing UNI foundation model achieved an AUC ranging from 0.73 to 0.93.

As shown in Table 2, model performance improved across the board when predicting the correct Bethesda category. UNI remained the top performer, with Top-1, Top-3, and Top-5 scores increasing to 0.75, 0.83, and 0.94, respectively. This improvement highlights the model's enhanced capability, particularly with RAG. augmentation—to assess malignancy likelihood independently of the specific diagnosis.

**Table 1.** Comparison of foundational models for predicting surgical diagnosis from thyroid cytology images.

| | Model Performance Predicting Surgical Diagnosis | | | | |
|---|---|---|---|---|---|
| | UNI | Gigapath | Virchow | ViT-32 | Ensemble |
| Top-1 | 0.69 | 0.56 | 0.50 | 0.25 | 0.42 |
| Top-3 | 0.81 | 0.61 | 0.58 | 0.31 | 0.56 |
| Top-5 | 0.92 | 0.64 | 0.61 | 0.33 | 0.64 |

**Table 2.** Comparison of foundational models for predicting TBSRTC from thyroid cytology images.

| | Model Performance Predicting TBSRTC | | | | |
|---|---|---|---|---|---|
| | UNI | Gigapath | Virchow | ViT-32 | Ensemble |
| Top-1 | 0.75 | 0.61 | 0.61 | 0.33 | 0.47 |
| Top-3 | 0.83 | 0.67 | 0.72 | 0.36 | 0.61 |
| Top-5 | 0.94 | 0.69 | 0.75 | 0.36 | 0.69 |

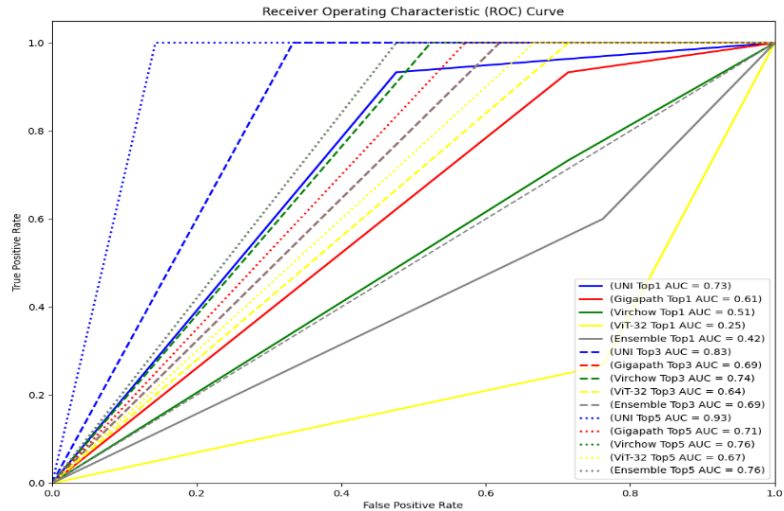

**Fig. 1.** Receiver operator characteristic curve of models for predicting surgical diagnosis. Top1 (solid line), Top3 (dashed line), and Top5 (dotty line) of models and their associated AUC are shown on the legend of the figure.

## 5   Discussion

In this study, we presented a comprehensive evaluation of various foundation models for predicting patient diagnoses from cytology samples. Our results, summarized in Table 1, demonstrate that pathology-specific models outperformed general-purpose vision transformers and ensemble approaches, with the UNI model achieving the highest accuracy. Our results demonstrate that pathology-specific models outperformed general-purpose approaches, with UNI achieving the highest accuracy (Top-1: 0.69, Top-3: 0.81, Top-5: 0.92). ROC and AUC analyses further confirm UNI's strong predictive performance (AUC: 0.73–0.93). Predicting Bethesda categories improved model performance overall, with UNI maintaining its lead (Top-1: 0.75, Top-3: 0.83, Top-5: 0.94). These findings highlight the advantages of pathology-specific models and retrieval-augmented generation (RAG) in enhancing cytology-based diagnostic predictions.

The Bethesda System for Reporting Thyroid Cytopathology (TBSRTC) recommends that no more than 7% of thyroid cytology cases be classified as Bethesda III, atypia of undetermined significance (AUS). In a clinical setting, this is challenging due to inter-observer variability among pathologists and the multiple follow up options available to patients.

Since its establishment in 2007, TBSRTC has been revised multiple times, the latest in 2023, to decrease ambiguity and aid pathologists in establishing a clear-cut diagnosis of thyroid cytology samples, particularly for intermediate categories Bethesda III and Bethesda IV. Despite these efforts, inter-observer disagreement remains significant at 10-40% between different observers. The revision of TBSRTC in 2023 has worked to decrease this disagreement with moderate success, highlighting the need for different approaches to tackle this ongoing issue [11].

For cases diagnosed as AUS with sufficient material for further analysis, follow-up approaches such as repeat FNAC biopsy, molecular testing, diagnostic lobectomy, or clinical surveillance are recommended based on the clinician's discretion [16]. However, repeat FNAC may not always be feasible due to a shortage of cytopathology specialists [31], and the uncertainty of an initial AUS diagnosis can lead to significant psychological distress for patients [32]. Additionally, the lack of absolute consensus on the recommended timeframe for a repeat FNAC complicates clinical decision-making [33].

Molecular testing has proven to be a valuable adjunct in reducing diagnostic ambiguity and improving surgical outcomes in AUS cases, but its cost remains a significant barrier to widespread use [34]. While molecular testing aids in risk stratification, diagnostic lobectomy remains the gold standard for post-operative diagnosis of thyroid malignancies, with paraffin block sample preparation providing definitive histopathological confirmation [35]. However, this procedure is also associated with additional costs and potential morbidity, particularly for patients who receive a negative result or opt for long-term surveillance [34]

We recognize that the sample size is relatively small and acknowledge the need for further validation of our findings in a larger study population. However, these initial findings lay a strong foundation for future research. To enhance reproducibility, the authors plan to expand this work with a larger dataset, though this is beyond the scope of the current study.

## 6   Conclusion

The challenges discussed highlight the need for tools that aid pathologists, surgeons, and clinicians in managing patients with unclear cytology diagnoses. The application of pathology foundation models combined with retrieval-augmented generation offers a promising outlook to fill in this gap as highlighted by the positive results of the experiment.

**Acknowledgments.** This study was internally funded by our institution.

**Disclosure of Interests.** The authors have no competing interests